\documentclass[conference]{IEEEtran}

\usepackage{mathtools}
\usepackage[caption=false, font=footnotesize]{subfig}
\usepackage{graphicx}
\usepackage{hyperref}
\usepackage{cleveref}
\usepackage{float}
\usepackage{booktabs}
\usepackage{verbatim}
\usepackage{chngcntr}

\newcommand{\xv}{\mathbf{x}}

\newcommand{\yv}{\mathbf{y}}

\newcommand{\wv}{\mathbf{w}}

\newcommand{\Bv}{\mathbf{B}}
\newcommand{\Nv}{\mathbf{N}}

\newcommand{\ra}[1]{\renewcommand{\arraystretch}{#1}}
\newcommand*{\specialcellbold}[2][b]{%
  \bfseries
  \begin{tabular}[#1]{@{}c@{}}#2\end{tabular}%
}

\title{Strategies for Robust Image Classification}

\author{
\IEEEauthorblockN{Tom E. Cavey}
\IEEEauthorblockA{Dept. Computer Science\\
Colorado State University\\
\href{mailto:tomcavey@colostate.edu}{tomcavey@colostate.edu}}
\and
\IEEEauthorblockN{Andy A. Dolan}
\IEEEauthorblockA{Dept. Computer Science\\
Colorado State University\\
\href{mailto:adolan5@colostate.edu}{adolan5@colostate.edu}}
\and
\IEEEauthorblockN{Jason D. Stock}
\IEEEauthorblockA{Dept. Computer Science\\
Colorado State University\\
\href{mailto:stock@colostate.edu}{stock@colostate.edu}}
}

\begin{document}
\maketitle
\thispagestyle{plain}
\pagestyle{plain}

\begin{abstract}

In this work we evaluate the impact of digitally altered images on the performance of artificial neural networks. We explore factors that negatively affect the ability of an image classification model to produce consistent and accurate results. A model's ability to classify is negatively influenced by alterations to images as a result of digital abnormalities or changes in the physical environment. The focus of this paper is to discover and replicate scenarios that modify the appearance of an image and evaluate them on state-of-the-art machine learning models. Our contributions present various training techniques that enhance a model’s ability to generalize and improve robustness against these alterations.

\end{abstract}

\section{Introduction}

Computer vision and image classification play primary roles in systems ranging from facial recognition to controlling autonomous vehicles. These technologies rely on machine learning models to produce results with a high classification accuracy. In a supervised environment, this accuracy is achieved by learning from prior experiences which allow for previously unseen input samples to be identified. 

A key challenge for training these models is the immense variety of scenarios that may be encountered by image classification systems in the real world. Additionally, images as they are captured may be subject to external forms of visual interference and distortion. These scenarios may be a deviation from an expected scene in a natural environment, or occur from physical hardware imperfections or image capture techniques. Nevertheless, the source of augmentation to a machine learning model is indistinguishable, but the outcome amplifies the risks of misclassification, leading to unintended or unsafe results.

\subsection{Environmental Perturbations}

Alterations in the physical environment that influence the input data into an image classifier may affect the performance of the network. For instance, a physical perturbation can occur when a transformation has altered the appearance of an object in the environment. Such augmentations may be naturally occurring while others are adversarial perturbations aimed to intentionally disrupt an image classifier. Disturbances within the environment such as graffiti, precipitation, or unforeseen objects are examples of physical environmental perturbations. 

A maliciously crafted sticker \cite{adversarial-patch} can be used to deceive image classifiers. A model will make false predictions based on the shape and position of these stickers within a scene. Evtimov {\it et al.} \cite{robustphysical} extend this example by showing how street signs can be modified with stickers that fool a model into classifying a stop sign as a speed limit sign. This differs from digital perturbations such that they occur independently of the camera hardware used to capture the input. Changes to the physical environment can be incredibly difficult to predict.

\subsection{Digital Perturbations}

Perturbations to an image may inevitably exist regardless of having a physically present augmentation in the scene. Situations as such can occur as a result of imperfections in the hardware or variations in the image capture technique. In either case, if the final image varies enough from the set of prior training samples, then it may fall outside the distribution of valid classes and be misclassified.

A change in the appearance of an image, whether significantly visible or not, can be captured with variations in the variables that control the exposure of a photograph: ISO, aperture, and shutter speed. Also known as the exposure triangle, these three variables require balance to correctly expose an image, as individually adjusting one will yield a change to the others. Increasing the ISO will proportionally increase the digital sensor's sensitivity to light but introduce noise. Using a wider aperture will allow more light to hit the sensor but decrease the depth of field and increase the possibility of undesired blur. Slowing the shutter speed will allow for more light into the camera over time but increase the possibility of undesired motion. 

The aforementioned scenarios are most commonly found in low light scenes where one or many of the variables are elevated to expose the image. However, even in a well lit environment with a proper exposure, there may exist irrecoverable hardware imperfections that result in slightly modified images. The most common occurrences are the introduction of dead, stuck, and hot pixels to the sensor of the camera. A dead pixel is permanently without power and appears black as all three sub-pixels remain unlit. Stuck pixels have one or more of the sub-pixels without power, resulting in a pixel that is red, green, blue, or any combination of these. A hot pixel will appear brighter than it is supposed to as a result of excess electrical charges increasing the voltage of a specific pixel.

\section{Related Work}

Previous work has been conducted to create adversarial images that aim to cause mislabeling by image classifiers. One paper that generally covers adversarial machine learning misclassification is \cite{goodfellowadversarial}, in which the susceptibility of neural networks to adversarial misclassifications is primarily attributed to their linear nature. Various techniques have been applied to the different augmentation domains described above. For digital augmentations, single pixel modifications, the introduction of image noise, and image blur are notable methods for altering appearance. For augmentations to the physical environment, a number of different techniques have been developed, most framed as potential malicious attacks that could be enacted against image classifiers.

In \cite{onepixelfoolingdnn}, Su, Vargas, and Sakurai implement a method to produce targeted misclassifications of the Kaggle CIFAR-10 dataset by changing only a single pixel of an image. This approach utilized a semi-black-box approach that observed previously trained classifier output probabilities and iteratively modified candidate perturbations through differential evolution to produce altered images with a relatively high success rate of targeted misclassification. Through this approach, seemingly important pixels were able to be found and modified to cause misclassifications. Furthermore, their work shows that randomly chosen pixels provide significant success rates. This result motivates our research of evaluating different training techniques with random pixel modifications.

In \cite{simpleblackboxpert}, Narodytska and Kasiviswanathan first applied a perturbation of a single pixel to input images in order to cause a misclassification. In doing so, they observed that even these small perturbations can be sufficient to create misclassifications, especially in lower resolution data sets. They refined this technique by generalizing the idea of a single “critical pixel” to a group or “critical set” of pixels to perturb, and utilized a greedy local-search algorithm to determine which specific pixels to perturb.

Several works have investigated adversarial techniques for causing misclassifications that augment more than a single or small number of pixels. In \cite{universaladvpert}, Moosavi-Dezfooli {\it et al.} create an algorithm based on the input distribution of natural images that attempts to find the best rate of misclassification while minimizing the required degree of image alteration. The latter part of their goal is notable in that they try to produce image alterations that are nearly invisible to the eye, while maximizing the rate at which those alterations cause misclassifications. In contrast to this previous work, our goal is to replicate and improve robustness against image augmentations that can occur naturally, as opposed to perturbations that are specifically designed to go unnoticed.

More in line with our work, Kurakin, Goodfellow, and Bengio investigated in \cite{adversarialexamplesworld} how well adversarial image perturbations translate to the physical world. To do so, the authors first applied different algorithms to generate adversarial image examples, then photographed the resulting images after printing them on paper. Interestingly, the authors found that perturbations that were more subtle were likely destroyed during this translation and therefore did not cause misclassifications at as significant a rate as when images were fed directly into the classifier. The authors also evaluated different image transformations, similar to what we experiment with in our work, for the purpose of evaluating how well adversarial examples translate when these transformations are applied. They specifically conduct experiments related to changes in contrast and brightness, Gaussian blur, Gaussian noise, and JPEG encoding. For which different adversarial perturbations carry over with different rates of success.

One other work that investigates naturally occurring image perturbations is \cite{cnnblur}, in which Vasiljevic, Chakrabarti, and Shakhnarovich evaluate the effect of optical blur on image classification models. Additionally, they attempt to utilize fine-tuning of a pretrained model and improve the robustness of the model as a result.

Another aspect of image perturbations leading to misclassification is that of alterations to the physical environment. In \cite{robustphysical}, Eykholt {\it et al.} manipulate street signs in the real world with stickers to cause targeted misclassifications. An   approach explored by Sharif {\it et al.} in \cite{Sharif_2019} involves the creation of special eyeglass frames meant to cause targeted misclassification of facial recognition models, another example of changing the physical environment for this purpose.

Our work focuses on alterations that occur when an image is captured, such as noise and blur. While related works focus on changes to the physical environment, our research is nonetheless applicable. The methods and results we provide can be used to build robustness against adversarial techniques.

\section{Methodology}

The purpose of this work is twofold: First, we implement a subset of the augmentation techniques discussed above and validate that image classifiers trained on natural images exhibit degraded performance when they are applied with augmented input data. Second, we apply different techniques of modified training using our augmented data and conduct experiments to determine which provide the greatest improvement on robustness of classifiers.

This work focuses on regular digital augmentations as they are described above because of the potential implications faced by manufacturers of products that rely on image classifiers. Said manufacturers’ devices are susceptible to imperfections which can lead to the natural occurrences of these augmentations. In other words, we prioritize the digital augmentations with the goal of producing classifiers that are robust to the augmentations that are common to all image processing devices.

Moreover, addressing physical augmentations provides a challenge of scale. Manipulations to the physical environment can take on a virtually infinite number of forms. While digital augmentations only involve changes to the encoded values of digital images, changes to the physical environment can introduce any combination of new information into the image that is captured. We therefore conclude that we are better equipped to directly address regular digital augmentations.

Experiments are conducted on two different datasets, MNIST \cite{MNIST} and CIFAR-10 \cite{CIFAR10}, to get a better understanding of the impact of augmentation. Both of these datasets have 10 unique classes, but CIFAR is composed of larger three channel RGB images in a more natural setting. This setting is different than MNIST in that the input space is significantly larger and takes more effort to fit a model. Therefore, prior to exploring the impacts of augmentation, a model is found for training natural images with high enough accuracy to confidently depict each class. We hypothesize that both the impact of augmentation, as well as the degree of robustness, will vary between the two due to the complexity of the data and their associated models.

\section{Network Structure}

It is important to construct a model with high accuracy prior to exploring the impact on accuracy with various augmentations. Convolutional neural networks were employed for the image classification tasks of this work. Our structures leverage PyTorch to allow for the construction of models with a dynamic number of hidden and convolutional layers. In the forward pass of such network, an $n \times n \times c$ image $I$ passes through the first convolutional layer with $u_1$ units having a $k_1 \times k_1$ kernel size with strides of $s_1$ as,

\begin{align*}
C_1(n, m) &= I * k_1 \\
&= \sum_i{\sum_j{I(n - i, m - j) \cdot k_1(i, j)}}
\end{align*}

\noindent
where $*$ denotes the convolution. The shape of this template, with $p$ zeros padding the image, is $z_1 \times z_1 \times u_1$, where $z_1 = \frac{\,( n + 2\times p_1 - k_1 )\,}{\,s_1 \,}+ 1$. By using the correct padding values it is possible to maintain the same output dimensions as the input, thereby allowing for deeper network structures. Thereafter, the output of each convolution is normalized with a $\mu = 0$ and $\sigma^2 = 1$, similar to how the data is initialized before training starts. This is done to improve gradient flow, regularization, and stabilization during training. Normalization is done for each of the $u$ units in the previous layer, hence the batching. Given the output $\xv_k$ from the $k^{\text{th}}$ unit, we normalize as,

\begin{align*}
    \mu_k &\leftarrow \frac{1}{N} \sum_{i=0}^N{\xv_{k_i}} \\
    \sigma^2_k &\leftarrow \frac{1}{N} \sum_{i=0}^N{(\xv_{k_i} - \mu_k)^2} \\
    \hat{\xv}_k &\leftarrow \frac{\xv_k - \mu_k}{\sqrt{\sigma^2_k + \epsilon}} \; 
    \text{ : } \epsilon = 1\mathrm{e}{-5}\\
\end{align*}

\noindent
We follow the normalization layer with the Rectified Linear Unit (ReLU) activation function as $g_{\,ReLU}(x) = max(0, x)$. Therefore, a transformation must occur on $\hat{\xv}_k$ to shift the distribution away from $\mu_k = 0$. If otherwise, half the outputs from $g_{\,ReLU}$ would be set to zero. A set of parameters $\gamma$ and $\beta$ are used to shift and scale the values and model the activation function with default values of 1 and zero respectively. These values are learned through back propagation with the batch normalization layer. Thus we obtain an output $\yv_k \leftarrow \gamma \xv_k + \beta \equiv \Bv\Nv_{\gamma, \beta}(\xv_k)$. This value proceeds through the activation function, preserving the same shape as the input from the previous layer, where it is then used as the input to the next layer.

Moreover, we reduce training variance and computational complexity by utilizing max pooling layers and dropouts throughout our network structure. The concept of max pooling is similar to a convolutional layer with defined kernels and strides. However, instead of computing a convolution with the input value, we down-sample the input by capturing the maximum value within the kernel and drop the rest. Using a stride of $s = 2$ we reduce the size in half. It can be seen that no weights are being learned in this layer. This is an effective method for emphasizing edged lines with the strongest weights. Dropout layers follow each max pooling with a probability $p = 0.20$, from a Bernoulli distribution, of randomly zeroing elements from the input tensor. Hinton {\it et al.} \cite{hinton2012improving} showed this technique is effective to further regularize and prevent co-adaptation of units. Our experiments show that including this dropout layer has a positive impact on test accuracy by reducing overfitting of the training data. 

\begin{figure}[!b]
	\centering
	\subfloat[\label{fig:mnist_network_structure}]{%
        \includegraphics[width=0.7\linewidth]{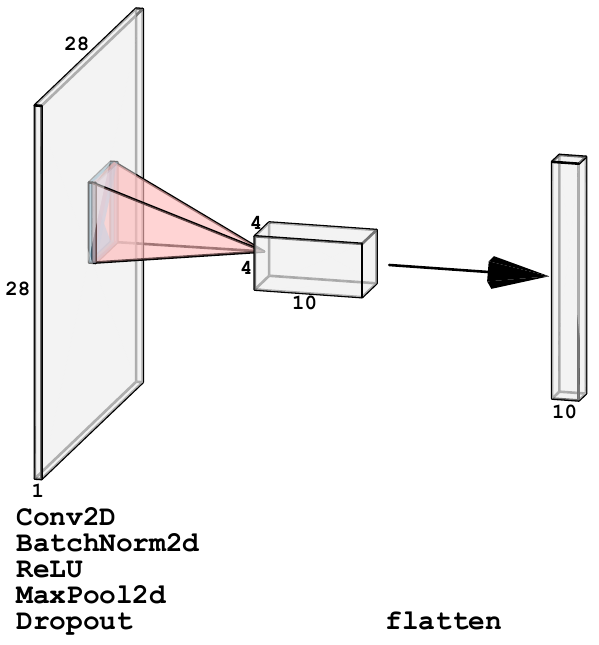}}
	\caption{Neural network structure trained for MNIST to achieve 97.420\% test accuracy. The model is trained with $\rho = 0.05$, over 50 epochs and a batch size of 1,500.}
\end{figure}

\begin{figure*}[!t]
	\centering
	\subfloat[\label{fig:cifar_network_structure}]{%
        \includegraphics[width=\textwidth]{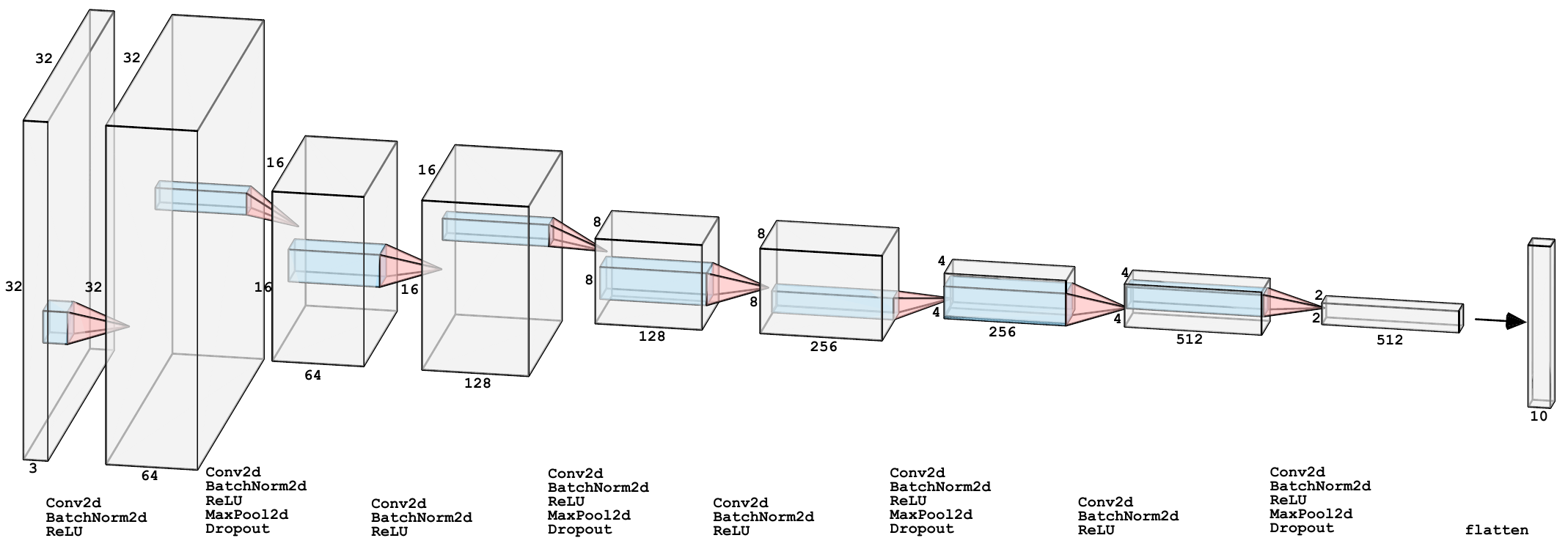}}
	\caption{Neural network structure trained for CIFAR-10 to achieve 86.920\% test accuracy. The model is trained with $\rho = 0.0005$, over 20 epochs and a batch size of 100.}
\end{figure*}

The inputs continue in the forward direction for $l$ convolutional layers, taking the output of the previous layer as input. The layers following are linear and require the output of the last convolutional layers to be flattened to a $u_l z_l^2$ vector. If there are fully connected layers, then for each unit there exists some $f(\xv_n, \wv_n)$ that is a function of $\xv_n$ which is parameterized by the subset of weights $\wv_n$. This is a simple linear transformation of the input, $g(\xv_n^T \wv_n) = g(w_0 + w_1 x_{n,1} + w_2 x_{n,2} + \cdots + w_D x_{n,D})$ where $w_0$ is multiplied by a bias value of 1. In the last layer, or if there are no fully connected layers, then the flattened output of $C_l$ passes into a linear output layer without $g_{\,ReLU}(x)$.
\newpage
From here, the model's output is calculated by computing the softmax on the output layer for each class $k$, denoted by,

\begin{align*}
\sigma_k(\xv_n)\;\;\;\;\; 
    \text{ where } \sigma(y_k) = \frac{e^{y_k}}{\sum_{m=1}^K e^{y_m}} \\
\end{align*}

This is used to represent the output $\yv$ as a probability distribution over all $k$ classes, where the maximum value is said to be the most probable class for some sample. Therefore, we choose to find weights $\wv$, in all layers, that maximizes the likelihood of data over all classes. This is denoted by the log likelihood function,

\begin{align*}
      LL(\wv) & = \sum_{n=1}^N \sum_{k=1}^K t_{n,k} \log \sigma_k(\xv_n) \\ 
\end{align*}

Since this function is differentiable, it can be used to update the weights using gradient descent. For each step, the function is optimized to minimize the loss using Adaptive Moment Estimation (Adam) to improve convergence and performance of training. We initialize Adam with a learning rate $\rho$ and beta parameters $\beta_n$ to tune how quickly and accurately the models will learn. Various values of $\rho$ are used to determine which contribution of the gradient is most effective for the data being used. Further, we specify each $\beta_n$ parameter as coefficients for computing running averages of the gradient and its square, which are initialized with $\beta_1 = 0.50$ and $\beta_2 = 0.999$.

Each dataset has a unique network structure in reference to the number of training epochs, batch size, number of layers and units, kernel size and strides, and interleaving of the standardization techniques described. The process of finding the appropriate structure can be strenuous, and by starting with shallow networks and minimal transformations, we can evaluate how increasing complexity impacts performance. Experimentation shows that MNIST requires a far less complicated structure to get high accuracy (\autoref{fig:mnist_network_structure}) as compared to that used to learn CIFAR (\autoref{fig:cifar_network_structure}). Therefore, we can observe the impact of structural attributes on performance with greater margins by evaluating contributions of modules when training the CIFAR model on natural images. Each of the conducted experiments utilized the same hyperparameters: training epochs, batch size, number of layers and units, and kernel size and strides. What varies is the inclusion and exclusion of each attribute, including: batch normalization, max pooling, dropouts, and various activation functions. The model that performed best leveraged all of the techniques and acted as a starting point for the evaluation of our augmentation techniques (\autoref{tab:network_attributes}).

\begin{table}[H]
  \caption{Comparison of accuracy on CIFAR-10 with differences of network structure components. All trials utilize the same hyperparameters.}
  \resizebox{1\columnwidth}{!}{%
  \ra{1}
  \begin{tabular}{@{}lcccc@{}}
   \toprule
    {\specialcellbold{BatchNorm2d}} &
    {\specialcellbold{Dropout}} &
    {\specialcellbold{Activation}} &
    {\specialcellbold{Train Acc}} &
    {\specialcellbold{Test Acc}} \\
   \midrule
    True  & True  & ReLU & 99.124 & 86.920 \\
    True  & False & ReLU & 98.110 & 84.520 \\
    False & True  & ReLU & 98.230 & 82.080 \\
    True  & True  & tanh & 98.278 & 81.570 \\
    False & False & ReLU & 97.882 & 80.010 \\
    False & False & tanh & 96.658 & 77.060 \\
   \bottomrule
  \end{tabular}
  }
  \label{tab:network_attributes}
\end{table}

\section{Impact of Perturbations}
\label{sec:impact_of_perturb}

\begin{figure*}[!t]
    \centering
    \subfloat[\label{fig:impact_visual_pixel}]{%
        \includegraphics[width=0.315\linewidth]{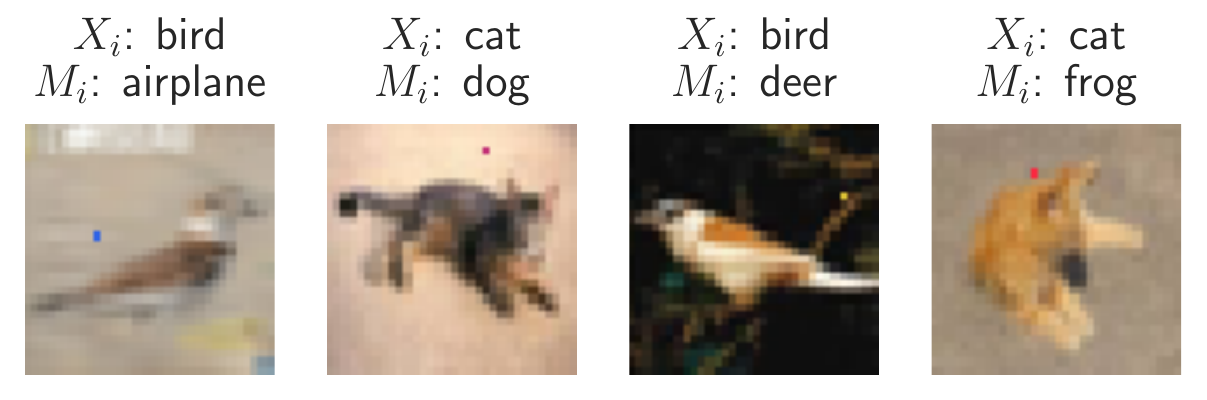}}\;
    \subfloat[\label{fig:impact_visual_noise}]{%
        \includegraphics[width=0.34\linewidth]{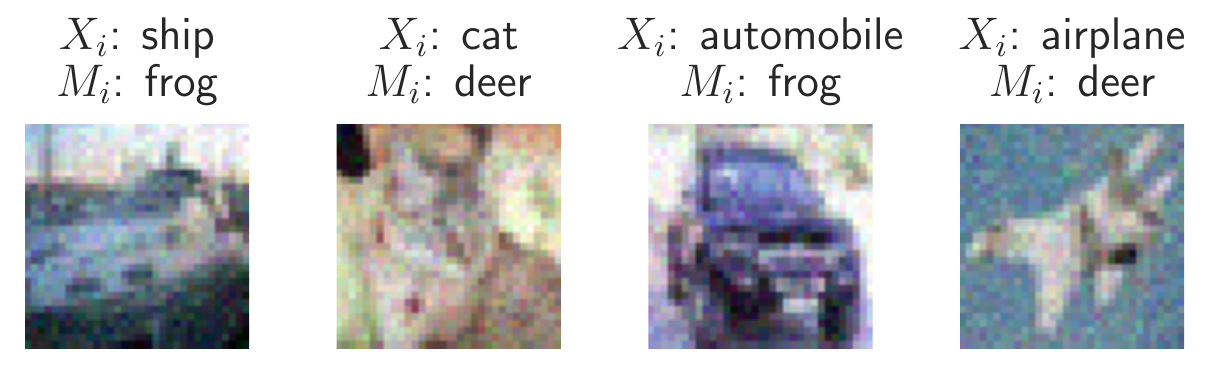}}\; 
    \subfloat[\label{fig:impact_visual_blur}]{%
        \includegraphics[width=0.315\linewidth]{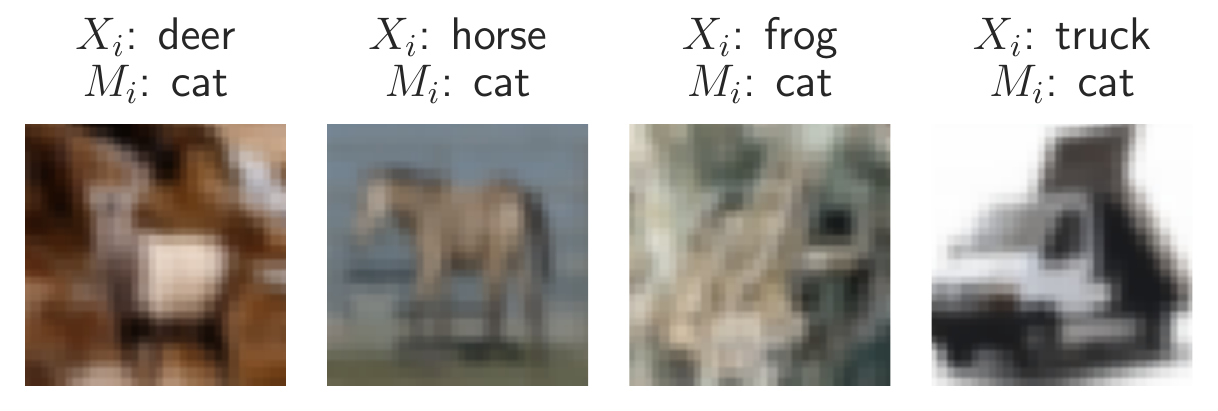}}\;
    \caption{$X_i$ represents the output of the trained model on the natural image, after applying the perturbation, the resulting $M_i$ prediction is shown. (a) displays the result of a single random stuck pixels, (b) shows a noise variance of 0.05, and (c) has added blur with a standard deviation for Gaussian kernel of 0.75.}
    \label{fig:impact_visual_all}
\end{figure*}

The impact that augmented data has on a model can be quantified as computing the percent correctly predicted with that data on a model trained on natural images. A percentage lower than that of the test accuracy of natural images signifies a degradation in performance. The test accuracy for MNIST and CIFAR yield 97.420\% and 86.920\%, respectively. Zagoruyko and Komodakis demonstrate an accuracy of 96.11\% is achievable on CIFAR-10 with Wide Residual Networks \cite{Zagoruyko2016wideresnet}. The difference in accuracy strongly correlates to the complexity of the input space and the models used. The focus of our work is to replicate and subsequently mitigate performance degradation caused by perturbed images.

Initial models trained on natural images have data ran with three main perturbations applied, including: modified pixels, image noise, and image blur. Each of these techniques have variations in the intensity of the corresponding modification. Results are captured and displayed for both MNIST and CIFAR datasets. We find the minimal tolerance for each technique, and scale the results to a meaningful yet reasonable limit. We discusse this topic in more depth below. \autoref{fig:impact_visual_all} shows a visual of the three perturbations types with the model's prediction on the natural and modified images. These results are studied in more detail via experiments to understand how variations of each impact accuracy.

\subsection{Pixel Modifications}

An image is represented as a $H \times W \times C $ multidimensional array where each $(H, W)$ index represents a pixel. All pixels within the image have a value between 0 and 1, inclusively, for each channel $C$. When $C > 1$, as seen with the RGB channels of CIFAR, the combinations of values yield specific color intensities. When modifying the intensities for a specific pixel, any of the channels may have their values changed.

The three variations for common pixel mutations in digital hardware include: stuck, dead and hot pixels. For each of these mutations we vary the number of pixels that are modified within the image to see if more pixels has a greater impact on accuracy. Modified pixels are randomly chosen from a discrete uniform distribution between $[0, W]$ for every experiment. However, to reduce the amount of variance in the results, we conduct a total of 25 trials for each variation in pixel values to obtain a mean and standard deviation and to increase precision. 

A stuck pixel can be simulated by setting each channel to a random value, from a continuous uniform distribution, with bounds between 0 and 1. As a result, the specified pixel intensity and hue has the ability to take on the appearance of the complementary, analogous, or any color range between. A hot pixel can be obtained by hard coding the pixel value for each channel to a value of 1, i.e., the maximum intensity. In contrast, a dead pixel is obtained by setting each channel value to a value of 0, i.e., minimum intensity for a specific pixel. 

\begin{figure}[H]
    \subfloat[MNIST\label{fig:impact_pixel_change_mnist}]{%
        \includegraphics[width=0.5\columnwidth]{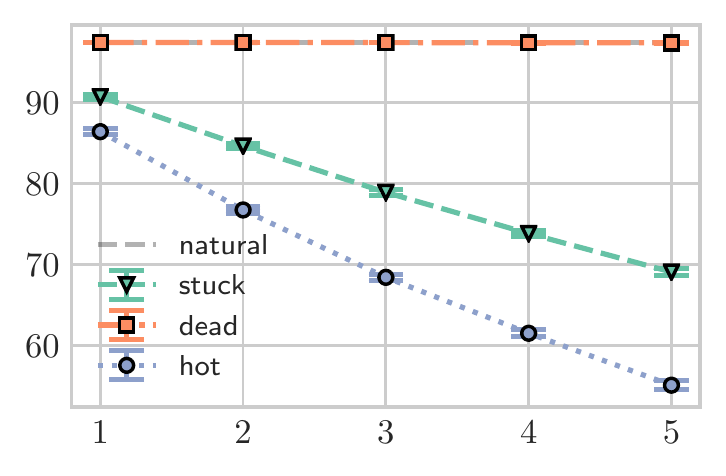}}
    \subfloat[CIFAR-10\label{fig:impact_pixel_change_cifar}]{%
        \includegraphics[width=0.5\columnwidth]{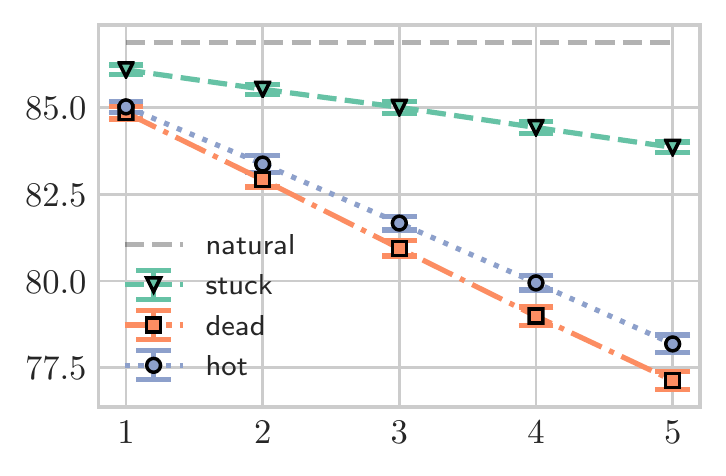}}
    \caption{(a) and (b) show the result of pixel perturbations of data on models trained on natural images. Test accuracy is shown as a function of the number of changed pixels.}
    \label{fig:impact_pixel_change}
\end{figure}

In theory, a camera may take on any number of pixel modifications, but their accumulated occurrence is not significantly high in practice. Therefore, we only explore how the discrete change of pixel values between 1 and 5 pixels influences accuracy. \autoref{fig:impact_pixel_change} illustrates the impact for each modification across the entire test set for the aforementioned number of trials on both MNIST and CIFAR datasets. Both experiments show a linear decrease in mean accuracy as the number of modified pixels increase for all two of the three described techniques. In \autoref{fig:impact_pixel_change_mnist} it can be observed that the occurrence of dead pixels has minimal impact on MNIST due to the majority of the image having a background value of zero. The reason for this is that for each convolution over an image the majority of the multiplications are zeroed out. Furthermore, the max pooling operation forwards the maximum, and as such, emphasizes the model's ability to recognize edges within the image. On the contrary, stuck and hot pixels drop accuracy roughly 30\% and 45\% respectively. This is due to the network being more adaptable by weighted entries in the image.

Dead pixel modifications in CIFAR do not carry the same traits seen in MNIST model as the data is significantly more heterogeneous. Instead, \autoref{fig:impact_pixel_change_cifar} shows the addition of stuck pixels having the least impact on performance. However, hot pixels being the most detrimental for CIFAR, only degrade accuracy by roughly 6\%. This result suggests that the input space allows for more variance and the model is inherently more robust to noise, but there is still room for improvement.

\subsection{Gaussian Noise}

In general, noise is related to some amount of random variance in a measurement. In the specific context of photography and digital images, noise relates to unwanted variance in the signal that is captured by an image sensor \cite{photoencyc}. It is the result of random variance in individual pixels that can be related to the electronics of the image sensor, its level of gain when capturing an image, or even the temperature of the sensor.

To simulate noise in images, we implemented a function to additively apply samples of augmentation values to the image sampled from a Gaussian distribution, parameterized by the distribution’s standard deviation: variance. In other words, every pixel in each input image was altered by some value from a normal distribution with a mean of 0. As the variance increases, the probability that a pixel is augmented by a more extreme value increases as well.

\begin{figure}[!b]
    \includegraphics[trim={0 0.85cm 0 0.7cm},clip,width=1\columnwidth]{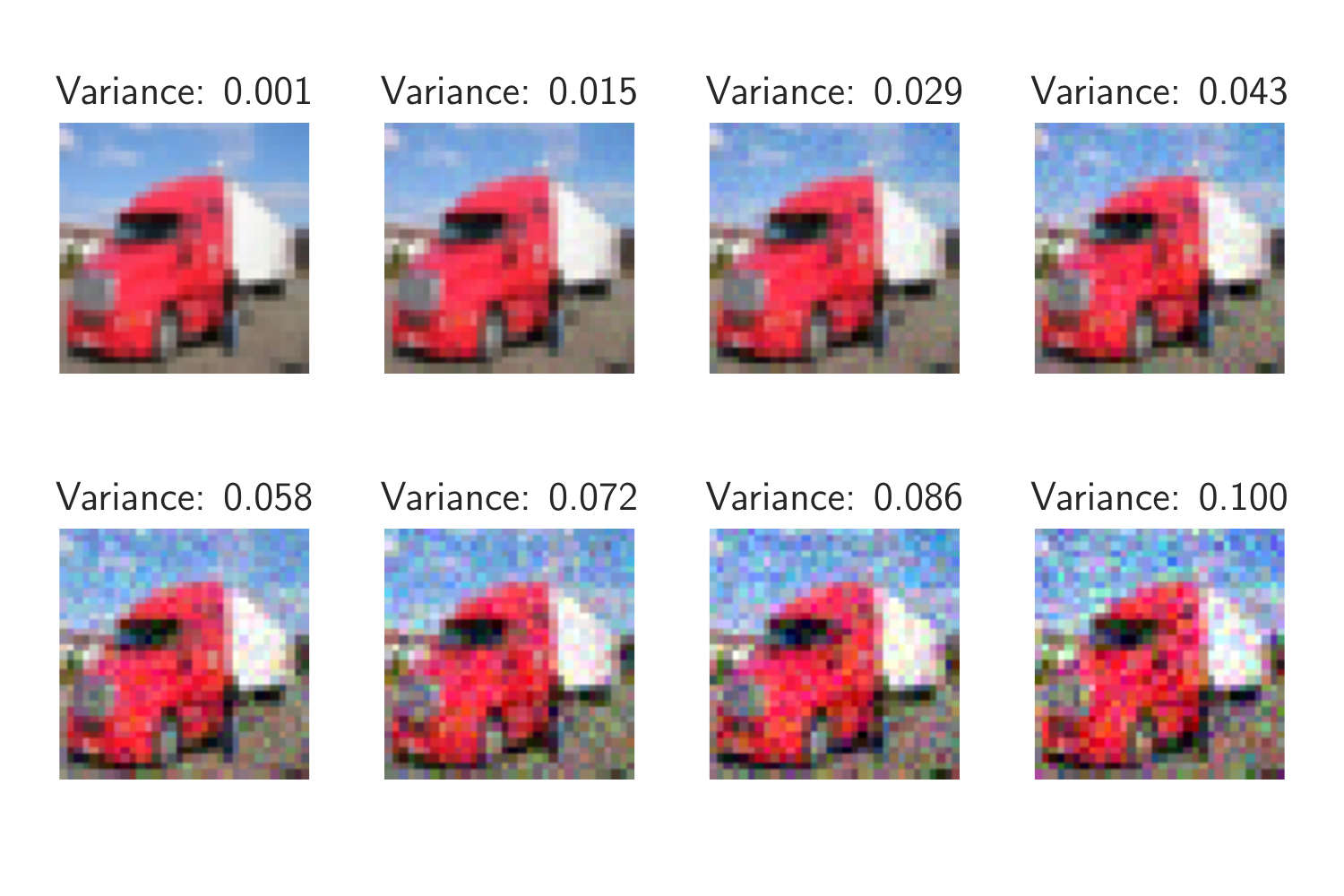}
    \caption{An image from CIFAR augmented with simulated noise sampled with different levels of variance, ranging from 0.001 to 0.1.}
    \label{fig:increasing_noise}
\end{figure}

To illustrate this process formally, we use the normal distribution denoted as $\mathcal{N} (0, \sigma^2)$, where the mean of the value change to any pixel is 0, and the variance $\sigma$ is an adjustable parameter that relates directly to the amount of noise in the resulting image. From this distribution, we sample values between -1 and 1. The higher the variance, the more often any one sample will be closer to -1 or 1. These samples are used to augment the image additively, as all pixels in the image are represented by a value between 0 and 1. The addition of these samples changes the values of every pixel in the image slightly, except for those changed by samples of exactly 0. If any resulting pixel is outside the range between 0 and 1, then the value is clipped accordingly. The end result is an augmented image that has its pixels changed with some visible degree of noise. Note that the samples for simulated noise augmentation are selected to match the dimensions of the corresponding data. More specifically, noise is applied to all three channels of CIFAR images, while it is only applied to the single channel of MNIST images.

We have found that a variance of 0.10 creates images with an unusually large amount of noise, while a variance of 0.05 results in a more realistic amount of noise in the images. \autoref{fig:increasing_noise} shows an example of different levels of noise applied to a single image. Considering these realistic views of image noise, we primarily evaluate and train models on data augmented with a variance of noise between 0.001 and 0.050.

\begin{figure}[H]
    \subfloat[MNIST]{\includegraphics[width=0.5\columnwidth]{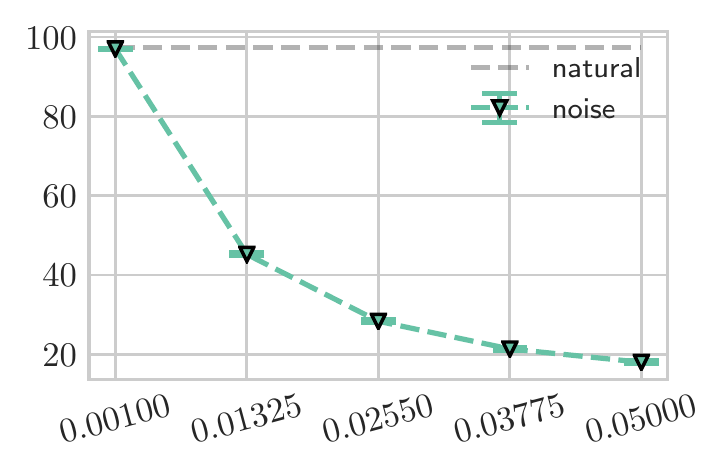}}
    \subfloat[CIFAR-10]{\includegraphics[width=0.5\columnwidth]{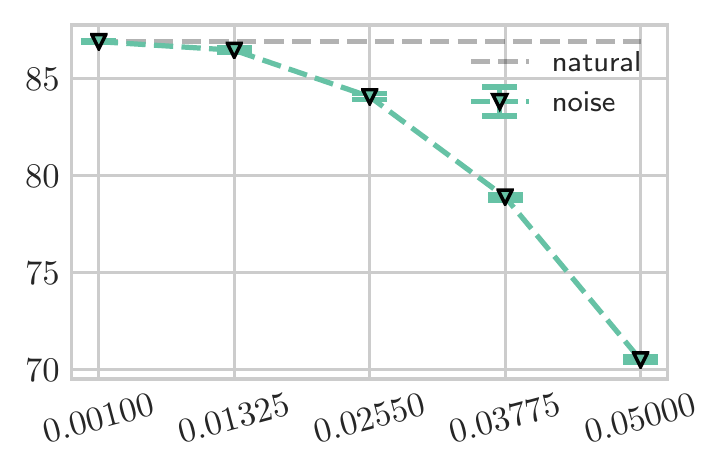}}
    \caption{(a) and (b) show the result of augmenting images with noise by change in variance, and testing on models trained on natural images. Test accuracy is shown as a function of the variance in noise.}
    \label{fig:impact_noise_change}
\end{figure}

When data augmented with simulated noise at different levels of variance was used as input and classified by the natural model, a clear degradation in classification accuracy was observed. This dropoff in performance related directly to the value of the variance used to sample the additive distribution of noise, clearly indicating that increasing amounts of noise introduced in the images resulted in worsening performance. \autoref{fig:impact_noise_change} illustrates the degree of this impact in performance for both MNIST and CIFAR. One notable aspect of the relationship between noise variance and model performance is the way that the MNIST model’s performance drops off much faster, and to much lower minimums, than the CIFAR model does. We attribute this difference to the nature of the different images. MNIST images consist of more homogeneous backgrounds that are more likely to be impacted by even slight alterations to pixels, while the images in the CIFAR dataset have more complex and heterogeneous backgrounds. The alteration of the background pixels therefore has more impact in disrupting the performance of the MNIST model, while the CIFAR model’s performance falls. Even so, the noise variance of 0.05 results in a classification accuracy of about 18\% for the MNIST model and about 63\% for the CIFAR model. The presence of noise in a digital image is common and occurs naturally, and these results motivate the need for improved robustness against such occurrences.

\subsection{Gaussian Blur}

An image may naturally have blur included for two primary reasons, (a) the image is out of focus due to the focal point of light being either in front or behind the camera sensor, or (b) by shooting with a wide aperture there is a decrease in the depth of field resulting in out of focus backgrounds. Simulating the latter is complex as the depth of an image has to be computed and blur applied outside the focal region of the photo. Whereas simulating (a) is a more intuitive process by artificially introducing a constant blur across the entire image, i.e., simulating the effect of moving the focal plane further from the focal point.

\begin{figure}[!t]
    \includegraphics[trim={0 0.85cm 0 0.7cm},clip,width=1\columnwidth]{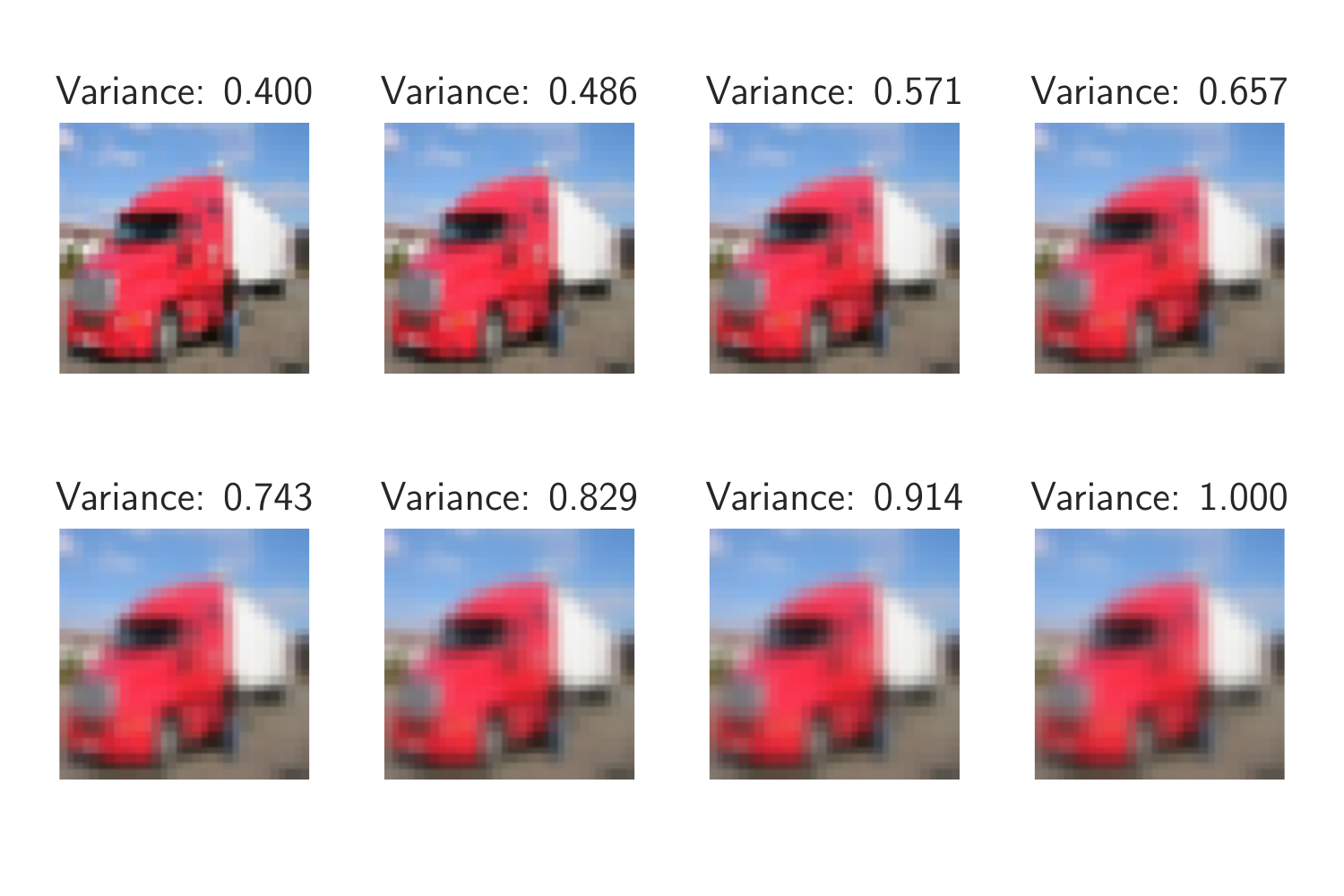}
    \caption{An image from CIFAR augmented with simulated blur sampled with different levels of variance, ranging from 0.04 to 1.00.}
    \label{fig:increasing_blur}
\end{figure}

Synthetic blur can be added to an image by scanning over every pixel in an image and recomputing their values based on a given pixel's surrounding values. This can be achieved by shifting a kernel over the image and performing a convolution for each pixel and channel in the image to ensure the dimensions do not change. This work utilizes Gaussian filtering to determine the weights within the kernel by approximating a 2D Gaussian with discrete values \cite{gaussian_filtering}. This method will ensure that central values have a higher weight as opposed to those on the periphery. An increase in the Gaussian standard deviation $\sigma$ will widen the bell curve, and influence a larger kernel, and add more blurring to the image.

In our experiments we vary the value of $\sigma$ between the range of 0.40 and 1.00. \autoref{fig:increasing_blur} displays these values with an example image from the CIFAR dataset. The output of applying the Gaussian filter is deterministic for a specified value of $\sigma$. Thus, we only have to apply the transformation once to the data to get a set of perturbed images. Given our original model trained on natural images, we test five different values of $\sigma$, linearly spaced between the aforementioned bounds, and record the resulting accuracies.

\begin{figure}[H]
    \subfloat[MNIST]{\includegraphics[width=0.5\columnwidth]{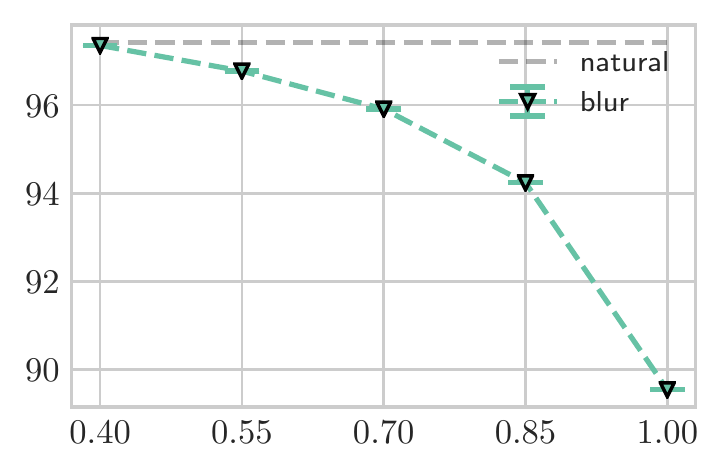}}
    \subfloat[CIFAR-10]{\includegraphics[width=0.5\columnwidth]{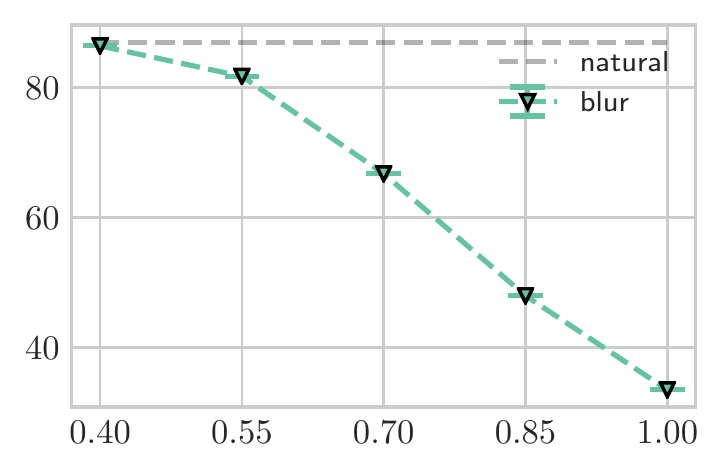}}
    \caption{(a) and (b) show the result of augmenting images with blur by change in variance, and testing on models trained on natural images. Test accuracy is shown as a function of the variance in blur.}
    \label{fig:impact_blur_change}
\end{figure}

\autoref{fig:impact_blur_change} displays the results for the MNIST and CIFAR models. Both have similar properties with a decrease in accuracy as $\sigma$ approaches 1. However, the results from MNIST only show a decrease in performance by 7\%. In contrast, the model for CIFAR shows an accuracy of 31\% when $\sigma=1$, i.e., a 55\% decrease in accuracy from natural images. This result is significant, but even minimal additions of blur can be detrimental to the model.

\section{Improving Robustness}
\label{sec:improving_robustness}

In this work we aim to create a model that provides high classification accuracy for these real world scenarios, explored in \autoref{sec:impact_of_perturb}, without being detrimental to the classification accuracy of natural images. The training techniques evaluated include: constant learning, incremental learning, and transfer learning. For each technique, input images are mutated with different combinations of alterations at varying degrees. Then the trained model is evaluated by testing how the range of alterations and natural images perform.

Exploring pixel modifications requires a combination of the type of mutation, i.e., dead, hot, and stuck, and number of these pixels changed for the training data with a range from 1 to 10 randomly selected pixels. Thereafter, the model is evaluated on each pixel type with two randomly selected pixels over 25 trials. The processes for noise and blur is similar, but require fewer models to be trained since there are less parameters that change. Specifically, with noise, there are five models trained at unique augmentations with linearly spaced variance between 0.001 and 0.050. Three augmented datasets, sampled over 25 trials, with 0.02550, 0.03775, and 0.05000 variance of applied noise are tested against these models (\autoref{fig:increasing_noise} illustrates a superset of these levels). The experiments with blurry images similarly have five model's trained on data with linearly spaced standard deviations of Gaussian blur applied between 0.04 and 1.00. These models are tested with three augmented datasets having blur applied at 0.500, 0.750, and 1.00 standard deviations.

We define robustness as a model's ability to have improved accuracy over applied perturbations while achieving an accuracy on the natural images that is similar to the original model. Therefore, we see an ideal situation when there is minimal variance in the accuracy for each test type and their grouping lies close to the original accuracy. In these experiments we show that the different training techniques, detailed in the following subsections, hold significant improvements over the range of scenarios.

\subsection{Constant Learning}

Constant learning is a simple method of training that involves only presenting perturbed images to a model. The focus of this approach is to make the augmentation appear insignificant to the network as every sample has some form of mutation applied to it. Data remains unaltered once training begins, and the model must learn to generalize the original data with a variety of modifications across samples. An immediate speculation may be that the trained model would exhibit poor performance on the natural images. However, our experiments show that this is not the case, and rather, accuracy of natural images remain high with greater improvements on perturbed images.

It can be observed in \autoref{fig:constant-pixel}, with the CIFAR dataset, that training on images modified with stuck pixels yields the most consistent results for each pixel type as compared to training on the other two types. Moreover, by training with two stuck pixels, accuracy for natural images is seen to be higher than the natural model with all types having a mean accuracy greater than 86\%, a 2\% improvement across all pixel types. However, the same cannot be said for training on hot or dead pixel training data. By training with hot pixels, a relative improvement is seen across all pixel types except for dead pixels. Conversely, by training on data with dead pixels, a degradation of performance by up to 4\% is seen with hot pixels. 

The results of MNIST hold significantly greater improvements when training with data perturbed by adding stuck or hot pixels. Specifically, a mean percentage greater than 97\% is seen for all types when training with data consisting of 7 hot pixels, a 20\% improvement over the worst case seen at two pixel changes in \autoref{fig:impact_pixel_change_mnist}. 

Prior to training with augmentations, the worst case performances from having images with noise or blur in the CIFAR dataset yield 63\% and 31\% respectively. However, by training models with the constant learning technique, we observe the prior worst case samples have a 21\% improvement on noise and 40\% improvement on blur. Accuracy for natural images and augmentation variances was maintained.

\subsection{Incremental Learning}

The incremental learning technique introduces perturbed images gradually into the dataset, increasing their quantity with each epoch during training. The neural network will be introduced to a small proportion of perturbed images during early stages of training and a large proportion of perturbed images at the final stages of training. Training begins with the dataset of images containing only 5\% of non-perturbed clean images based on the batch size. A ratio is calculated based on the epoch iteration count and the batch size. The algorithm determines an appropriate number of perturbed images based on this ratio and slices the clean and perturbed datasets at a given index. It can be hypothesized that an incremental training technique will help improve the overall robustness of the model by exposing the network to both clean and perturbed images simultaneously.

The experiments show that the robustness of the network is noteworthy, allowing the incremental learning to perform well on both clean and perturbed images. It can be observed in the following plot that the MNIST data with hot pixels appear to be in lock step and each perturbation within less than 0.5\% of other perturbations and natural image accuracy \autoref{fig:incremental-mnist-hot}. This is demonstrating that the incremental addition of perturbed images into the clean dataset does not significantly affect the model’s ability to accurately predict clean images.

The worst case performance of a model trained with clean images and tested on the CIFAR dataset with blur perturbation sits at roughly 31\%. By utilizing the incremental training technique, the model can be observed to improve worst case performance by 50\% on blurred images \autoref{fig:incremental-cifar-blur}. Additionally, it can be seen with the CIFAR data that the model trained on the highest amount of blur generally outperforms other models trained with less amounts of blur.

By far the most staggering result showcasing the effectiveness of incremental learning can be observed in the MNIST dataset with the noise perturbation. The performance of a model trained with clean images and tested on images perturbed with a noise variance of 0.05 is at roughly 19\%. By utilizing the incremental training technique the model can be observed to improve the model performance by 75\% on noisy images at 0.05 \autoref{fig:incremental-mnist-noise}.

\subsection{Transfer Learning}

The goal of transfer learning is to exploit the knowledge of one task, encoded in a previously trained network, for another task. The pretrained network's knowledge is utilized as a feature extractor, producing general or high-level abstract information from the data before additional new layers are trained on data for the specifics of the new task. In short, a network created for transfer learning may contain part or all of a pretrained network followed by a new network that is trained on new task data.

There are two approaches for how to handle the pretrained model's existing weights when adapting to the new task through additional training. One option is to freeze all parameters of the existing network and only allow back propagation to affect the parameters of the subsequent layers of the new task. The other option is to allow back propagation through both parts of the network, fine-tuning the parameters of the pretrained model in addition to the parameters of the new task’s layers.

\begin{figure}[!t]
	\centering
	\subfloat[\label{fig:transfer_learn}]{%
        \includegraphics[width=1\linewidth]{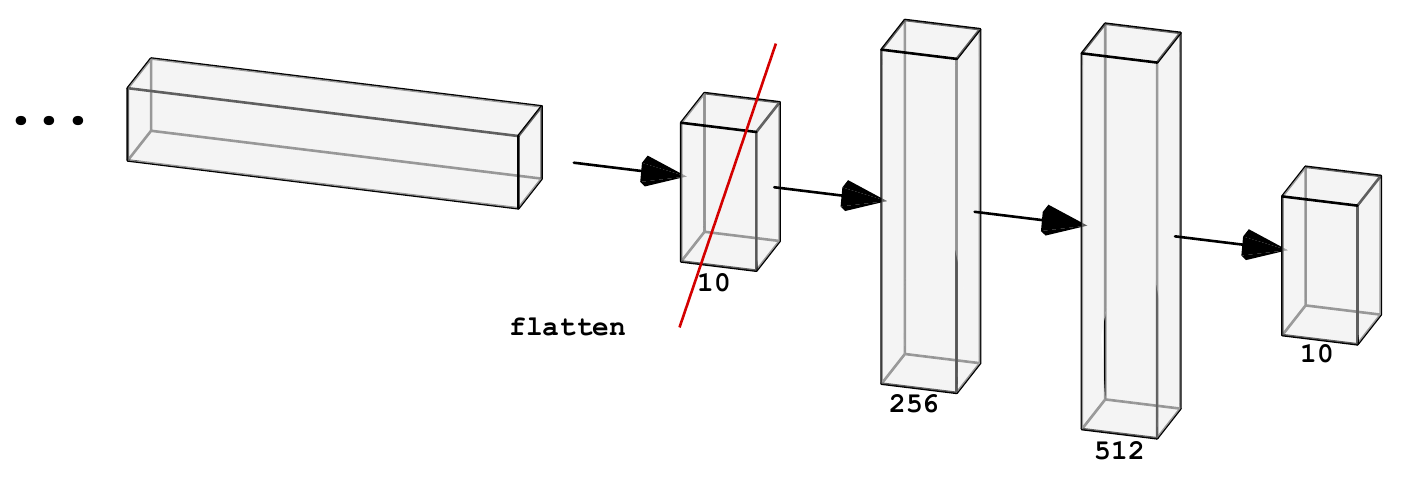}}
	\caption{Modified CIFAR model used for transfer learning, where the final output layer is replaced by two additional fully connected layers leading up to the final output layer.}
\end{figure}

From the perspective of this work, the “new task” for transfer learning is seemingly simple when compared to the original task that produces the pretrained model. All input images are the same, save for the augmentations that are applied to them described in \autoref{sec:impact_of_perturb}. Accordingly, our transfer learning networks for both MNIST and CIFAR are relatively simple in structure. In each model, the final output is produced by a linear layer which takes the flattened values after all convolutions and produces the probabilities for each class. For our transfer learning approach, we replace this final output layer with a new set of layers. For MNIST we append a fully connected layer with 256 units followed by the ReLU activation function before the final output layer. We do the same for CIFAR but with an additional fully connected layer of 512 units following the layer with 256 before the final output layer (\autoref{fig:transfer_learn}). The original models were pretrained on natural data from their respective datasets before the transfer learning models were constructed, thereby encoding the previous task’s weights in the new network. The training step for transfer learning involved training the newly modified network with augmented data. The final modified network is then used for evaluation.

In our experiments, we determined that allowing values to back propagate through the entire network provided slightly better results overall. The models chosen for each dataset resulted in the highest classification accuracies for noise-augmented images, the results of which are displayed in \autoref{fig:transfer-noise}. Specifically, while the pretrained model quickly fell to a classification accuracy of less than 65\% when evaluating noise-augmented images with a variance of 0.05, the CIFAR model trained with fine-tuned transfer learning on noise data of 0.05 variance maintained a classification accuracy of around 85\% on data of different variance values, including the natural images. This result shows that our implementation of transfer learning is a viable solution for building robustness against image noise perturbations.

While this structure was trained primarily on the noise-augmented images, it was also shown to perform with improved accuracy when trained and evaluated on data altered with other perturbation methods. For the pixel-related perturbations, while the CIFAR natural model’s classification accuracy fell as far as 2\% with two pixels changed, transfer learning with between 1 and 8 altered pixels significantly mitigated this performance dropoff. For the CIFAR model, the hot pixel perturbation most benefited from transfer learning, keeping the classification accuracy on hot-pixel data above 87\% as seen in \autoref{fig:transfer-pixel-cifar-hot}. One notable feature presented with this particular model is that its worst overall performance was with data containing dead pixels, though it still performed well when compared to the original model. This relationship follows intuitively, since the value of a dead pixel is the direct opposite of hot. As a result, the model is better adapted to hot pixels than dead pixels.

\section{Experimental Evaluation}

Results captured in \autoref{sec:improving_robustness} show that various training techniques can be employed to improve the robustness against a specific perturbation. These experiments resulted in a total of 240 models trained under varying conditions. Not every one performed as expected, but their quantity gave strong insights into how the different training techniques behaved. Specifically, we can observe best and worst case results for each perturbation type.

The presence of a modified pixel in an image will degrade performance against a natural model, but it exhibits the least amount of change among the three perturbation types. In \autoref{sec:impact_of_perturb} it was shown that pixels have at most 19\% and 2\% difference in accuracy with two pixel changes for MNIST and CIFAR respectively. Transfer learning was able to provide consistent improvements with accuracies deviating at most 1\% from their ground truth accuracy of natural images on the natural model. This reduces the number of misclassified samples from CIFAR in half, and shows MNIST dropoff being mitigated completely for hot and stuck pixels. Dead pixels show only marginal improvements regardless of training technique.

An image augmented with noise, for the worst case, is classified correctly 18\% and 62\% of the time for MNIST and CIFAR respectively. By utilizing incremental learning, we can see the greatest improvement for CIFAR by bringing all tested variances within 1\% of the ground truth. The other two techniques do not show the same improvements. However, all three are proven to be effective methods against MNIST by increasing the rate of classifications to 1\% of ground truth.

Improvements to classifying images perturbed with blur are most prominent with transfer and incremental learning techniques. We find that training models with a high amount of blur is most effective for obtaining high accuracies on natural and all variations of blur in images. Incremental learning shows that training with our upper bound where $\sigma=1$, we get results within 3\% of ground truth for CIFAR, an improvement of 60\%. Transfer learning provides the best results for the MNIST model with results within 1\% of ground truth.

Incremental and transfer learning have the greatest impact on performance improvements across the three perturbation types. Constant learning, with the most simplistic training model shows notable improvements but fails to triumph the other training methods.

\section{Insights}

Throughout the development of this project we encountered opportunities to improve the time and compute complexities, improve upon the machine learning model’s logic principles, and create automation tools to aide in the ease of training, testing, and visualization.

During the discovery phase of finding network structures with high accuracy performance, an observation was made that models were overfitting the training data. For example, a given model would perform at above 99\% accuracy when classifying images in the training dataset and 65\% on images in the testing data set. We hypothesized that if the proper methods are used to reduce overfitting, our testing accuracy can be increased by a substantial amount. We were able to reduce overfitting and improve the training process by implementing dropout, max pooling, and batch normalization methodologies.

Experimenting with the CIFAR dataset introduced a multitude of challenges tied to time and memory constraints. Early on, this hindered our ability to thoroughly explore many network structures in a timely manner. A solution we found beneficial was to train data in batches to reduce memory consumption and allow for the use of deep neural networks.

\section{Conclusions}

In this work we show that a small modification in the intensity of a single pixel or the presence of noise or blur in an image causes significant performance degradation in models trained on classifying unaltered images. In the worst case, the incorporation of such manipulations shows a drop in accuracy of over 80\%, yielding results that are no better than a random guess. We find this is due to the models' inability to generalize from their natural training data. This impact is significant because each perturbation type may naturally appear with prolonged hardware usage of cameras and slight deviations in the exposure triangle. 

Our analysis explores various training techniques to improve the accuracy over applied perturbations while achieving high accuracy on the natural images. We contrast the impact of constant, incremental, and transfer learning methods across a combination of scenarios, thereby improving a model’s ability to recognize perturbed images. In particular, we showed that leveraging these specific techniques results in an accuracy within 3\% of the natural model for all perturbations. In future work, we plan to explore the boundaries for the distribution of misclassified samples to understand how classes are confused. Furthermore, we would like to evaluate how the combination of training techniques and perturbation types can be utilized to create a globally generalized model that is robust to any and all of the evaluated perturbations.

\bibliographystyle{IEEEtran}
\bibliography{refs.bib}

%% -------------------------------------------------------------------- %%
%%                          START APPENDIX                              %%
%% -------------------------------------------------------------------- %%

\newpage
\onecolumn
\appendix

\counterwithin{figure}{subsection}
\setcounter{figure}{0}

\subsection{Constant Learning}

\begin{figure}[H]
    \centering
    \textbf{MNIST\hspace{12em}CIFAR-10}\par\medskip
    
    \subfloat[Stuck\label{fig:constant-mnist-stuck}]{%
        \includegraphics[width=0.30\textwidth]{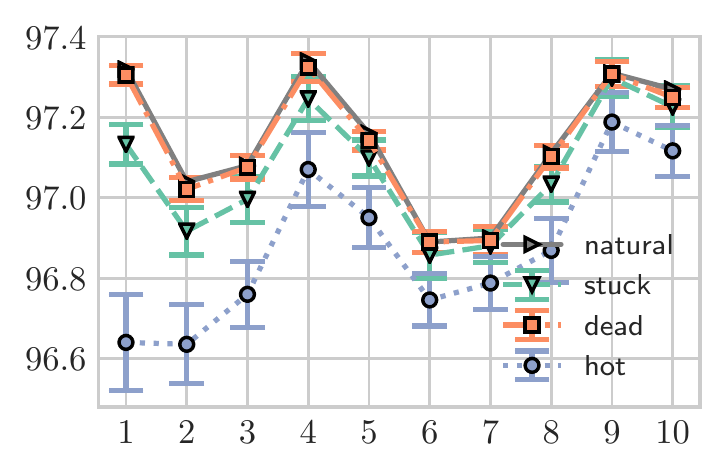}}\;
    \subfloat[Stuck\label{fig:constant-pixel-cifar-stuck}]{%
        \includegraphics[width=0.30\textwidth]{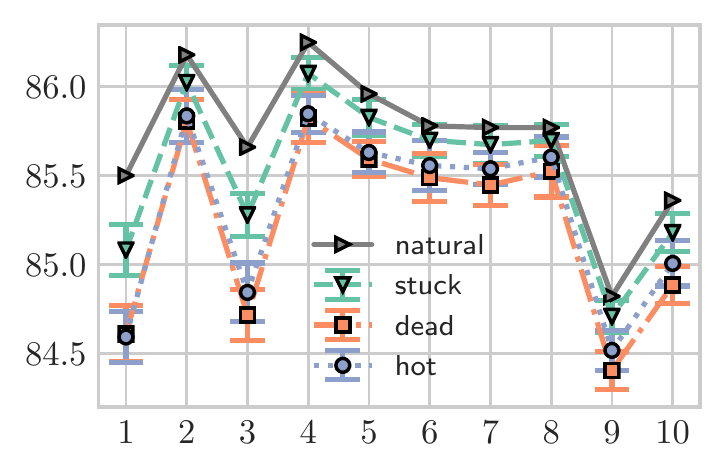}}
    
\subfloat[Hot\label{fig:constant-mnist-hot}]{%
        \includegraphics[width=0.30\textwidth]{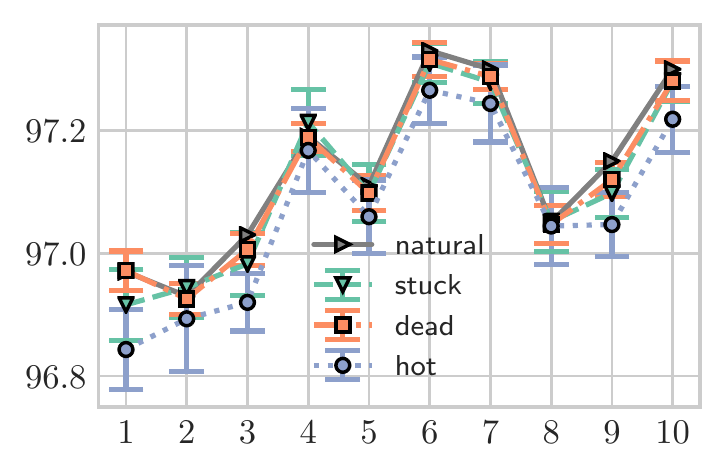}}\;
    \subfloat[Hot\label{fig:constant-pixel-cifar-hot}]{%
        \includegraphics[width=0.30\textwidth]{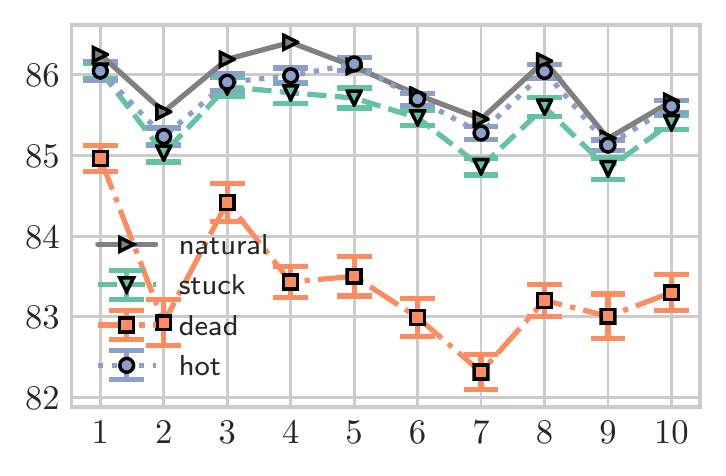}}
        
    \subfloat[Dead\label{fig:constant-mnist-dead}]{%
        \includegraphics[width=0.30\textwidth]{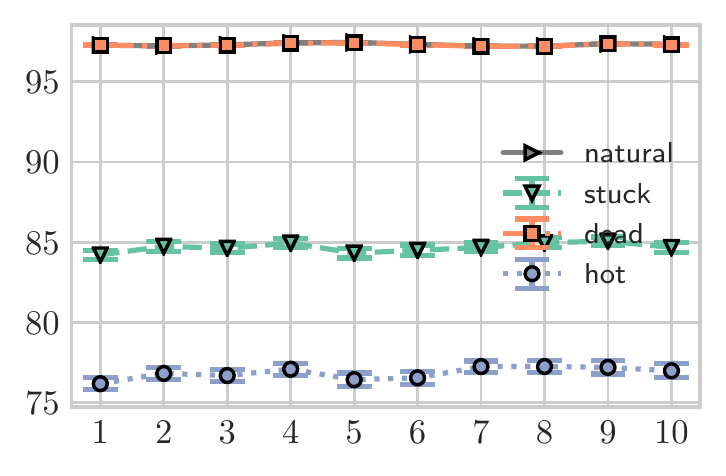}}\;
    \subfloat[Dead\label{fig:constant-pixel-cifar-dead}]{%
        \includegraphics[width=0.30\textwidth]{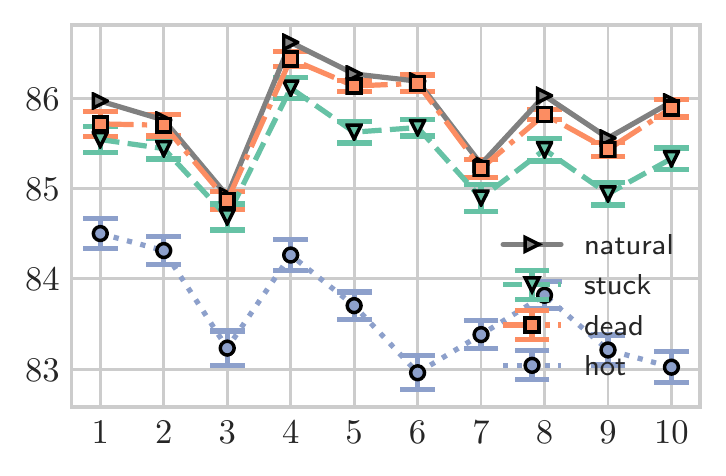}}
        
    \begin{minipage}{12cm}
    \caption{Constant learning models trained on pixel variations for each pixel category. Results show tests on 2 px changes sampled 25 times for the MNIST and CIFAR-10 datasets. Test accuracy is shown as a function of the change in the number of training pixels from 1-10.}
    \end{minipage}
    \label{fig:constant-pixel}
\end{figure}
    
\begin{figure}[H]
    \centering
    \textbf{MNIST\hspace{12em}CIFAR-10}\par\medskip
    
    \subfloat[\label{fig:constant-mnist-noise}]{%
        \includegraphics[width=0.30\textwidth]{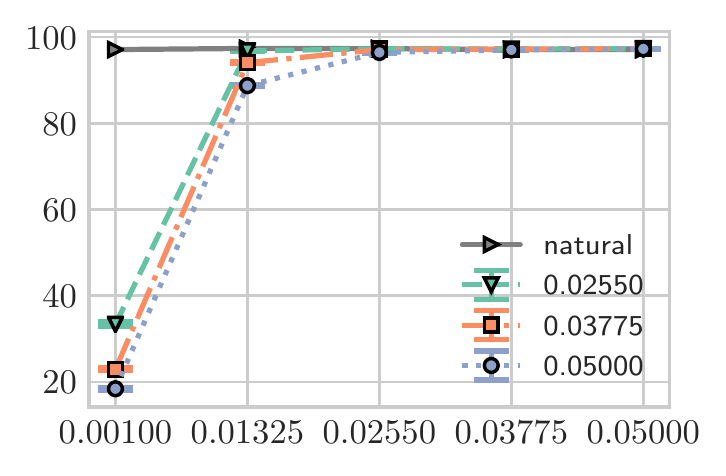}}\;
    \subfloat[\label{fig:constant-cifar-noise}]{%
        \includegraphics[width=0.30\textwidth]{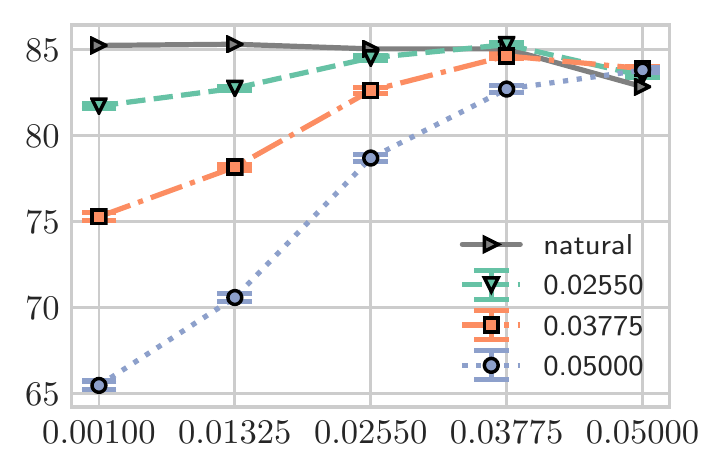}}
        
    \begin{minipage}{12cm}
    \caption{Constant learning models trained on variations of noise sampled from a normal distribution. Results show tests on various noise levels for the MNIST and CIFAR-10 datasets. Test accuracy is shown as a function of the variance in noise.}
    \end{minipage}
    \label{fig:constant-noise}
\end{figure}

\begin{figure}[H]
    \centering
    \textbf{MNIST\hspace{12em}CIFAR-10}\par\medskip
    
    \subfloat[\label{fig:constant-mnist-blur}]{%
        \includegraphics[width=0.30\textwidth]{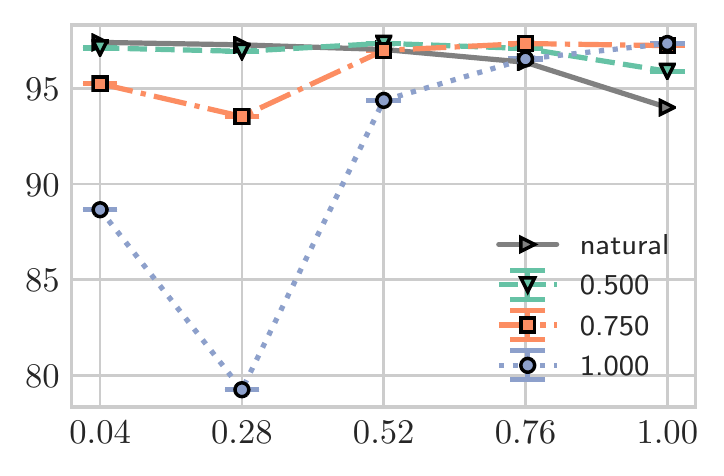}}\;
    \subfloat[\label{fig:constant-cifar-blur}]{%
        \includegraphics[width=0.30\textwidth]{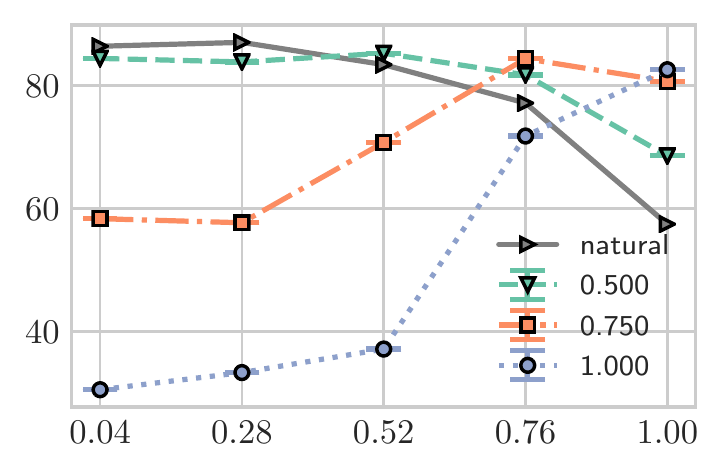}}
        
    \begin{minipage}{12cm}
    \caption{Impact of blurry input images applied to MNIST and CIFAR-10 models trained with constant learning. Test accuracy is shown as a function of the variance in blur.} 
    \end{minipage}
    \label{fig:constant-blur}
\end{figure}

\subsection{Incremental Learning}

\begin{figure}[H]
    \centering
    \textbf{MNIST\hspace{12em}CIFAR-10}\par\medskip
    
    \subfloat[Stuck\label{fig:incremental-mnist-stuck}]{%
        \includegraphics[width=0.30\textwidth]{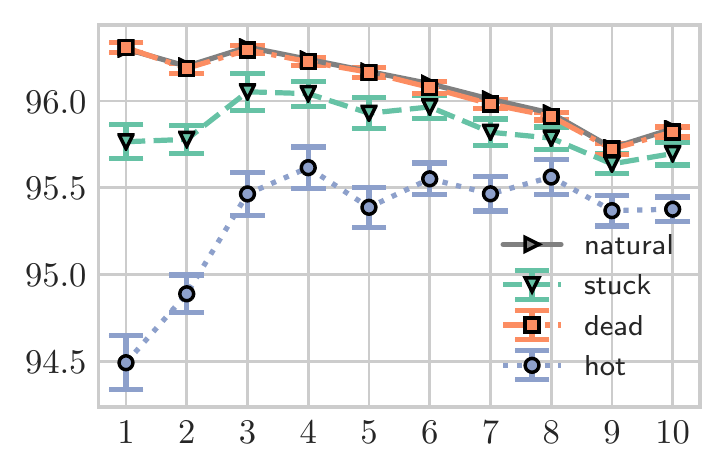}}\;
    \subfloat[Stuck\label{fig:incremental-cifar-stuck}]{%
        \includegraphics[width=0.30\textwidth]{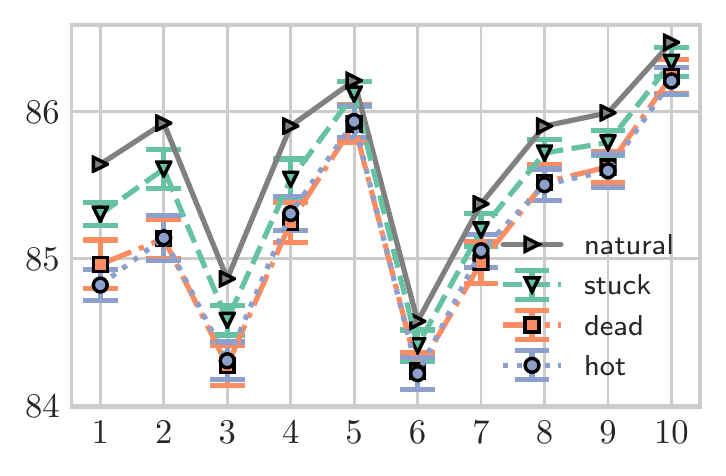}}
    
    \subfloat[Hot\label{fig:incremental-mnist-hot}]{%
        \includegraphics[width=0.30\textwidth]{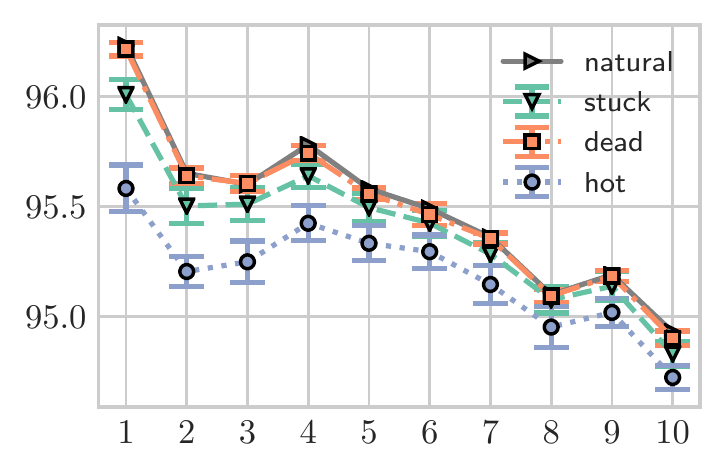}}\;
    \subfloat[Hot\label{fig:incremental-pixel-cifar-hot}]{%
        \includegraphics[width=0.30\textwidth]{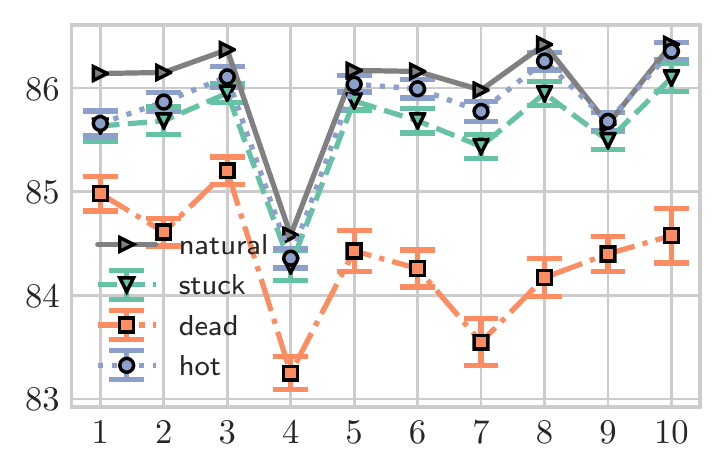}}
        
    \subfloat[Dead\label{fig:incremental-mnist-dead}]{%
        \includegraphics[width=0.30\textwidth]{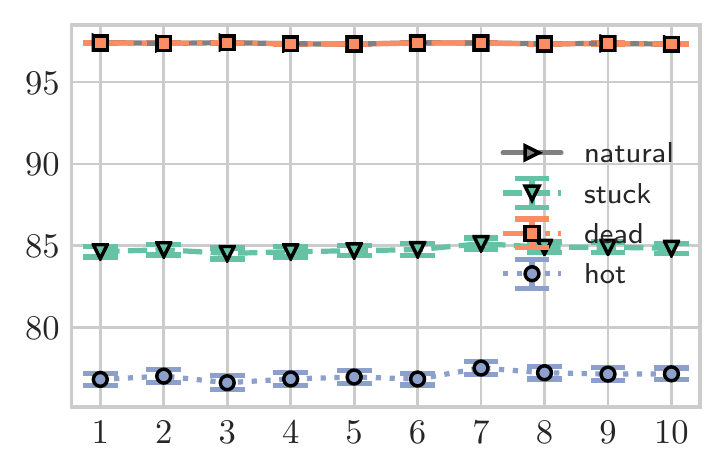}}\;
    \subfloat[Dead\label{fig:incremental-pixel-cifar-dead}]{%
        \includegraphics[width=0.30\textwidth]{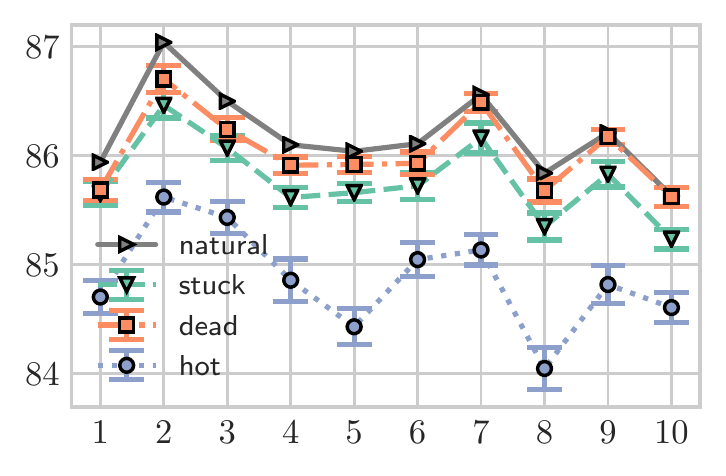}}
        
    \begin{minipage}{12cm}
    \caption{Incremental learning models trained on pixel variations for each pixel category. Results show tests on 2 px changes sampled 25 times for the MNIST and CIFAR-10 datasets. Test accuracy is shown as a function of the change in the number of training pixels from 1-10.} 
    \end{minipage}
    \label{fig:incremental-pixel}
\end{figure}

\begin{figure}[H]
    \centering
    \textbf{MNIST\hspace{12em}CIFAR-10}\par\medskip
    
    \subfloat[\label{fig:incremental-mnist-noise}]{%
        \includegraphics[width=0.30\textwidth]{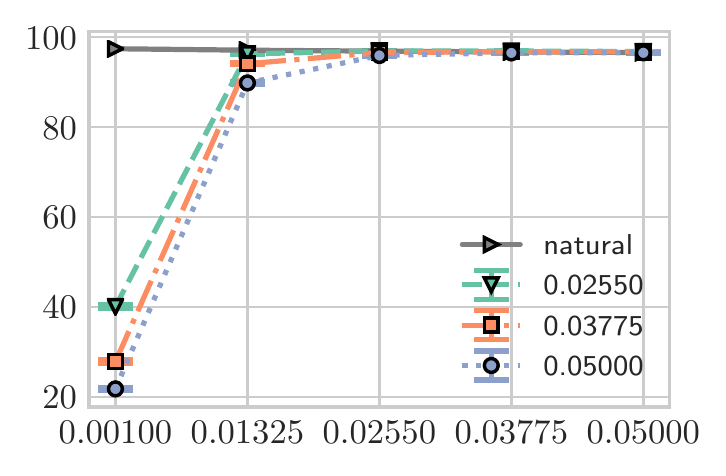}}\;
    \subfloat[\label{fig:incremental-cifar-noise}]{%
        \includegraphics[width=0.30\textwidth]{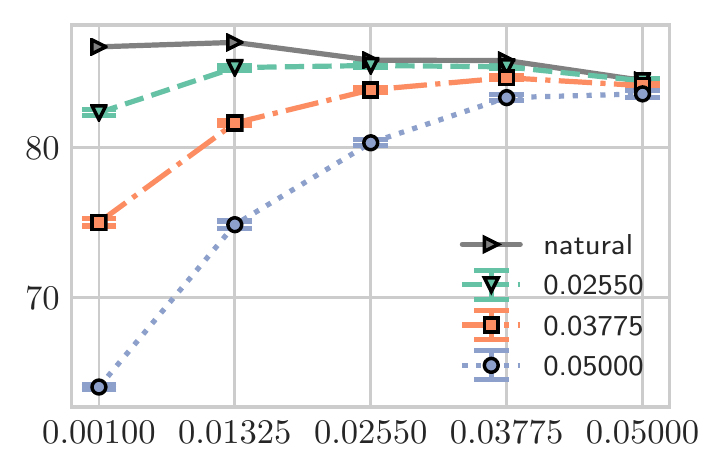}}
        
    \begin{minipage}{12cm}
    \caption{Incremental learning models trained on variations of noise sampled from a normal distribution. Results show tests on various noise levels for the MNIST and CIFAR-10 datasets. Test accuracy is shown as a function of the variance in noise.} 
    \end{minipage}
    \label{fig:incremental-noise}
\end{figure}

\begin{figure}[H]
    \centering
    \textbf{MNIST\hspace{12em}CIFAR-10}\par\medskip
    
    \subfloat[\label{fig:incremental-mnist-blur}]{%
        \includegraphics[width=0.30\textwidth]{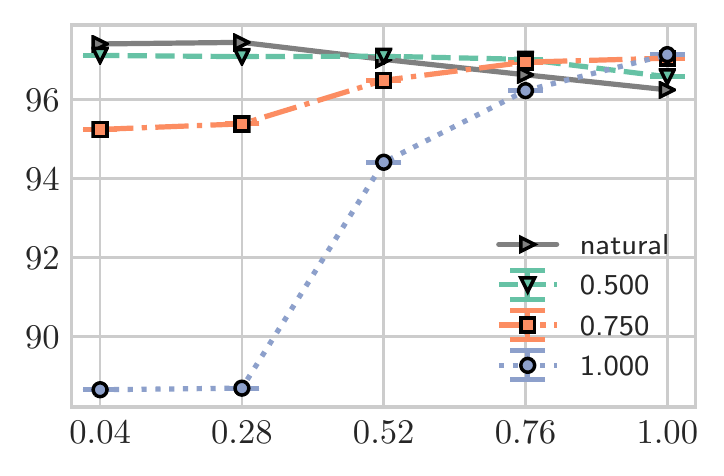}}\;
    \subfloat[\label{fig:incremental-cifar-blur}]{%
        \includegraphics[width=0.30\textwidth]{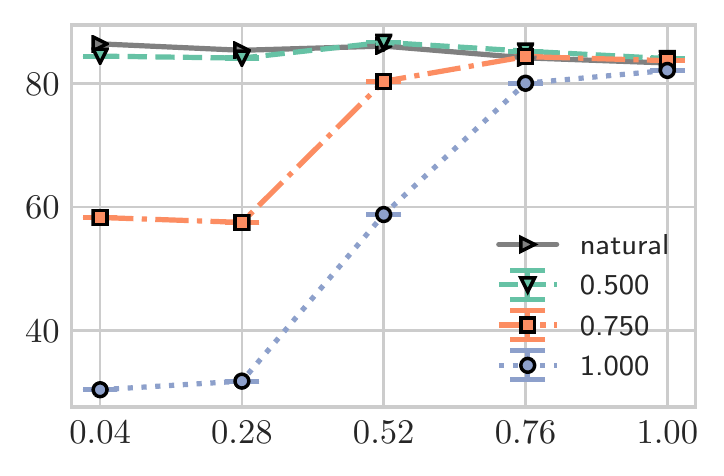}}
        
    \begin{minipage}{12cm}
    \caption{Impact of blurry input images applied to MNIST and CIFAR-10 models trained with incremental learning. Test accuracy is shown as a function of the variance in blur.} 
    \end{minipage}
    \label{fig:incremental-blur}
\end{figure}

\newpage

\subsection{Transfer Learning}

\begin{figure}[H]
    \centering
    \textbf{MNIST\hspace{12em}CIFAR-10}\par\medskip
    
    \subfloat[Stuck\label{fig:transfer-mnist-stuck}]{%
        \includegraphics[width=0.30\textwidth]{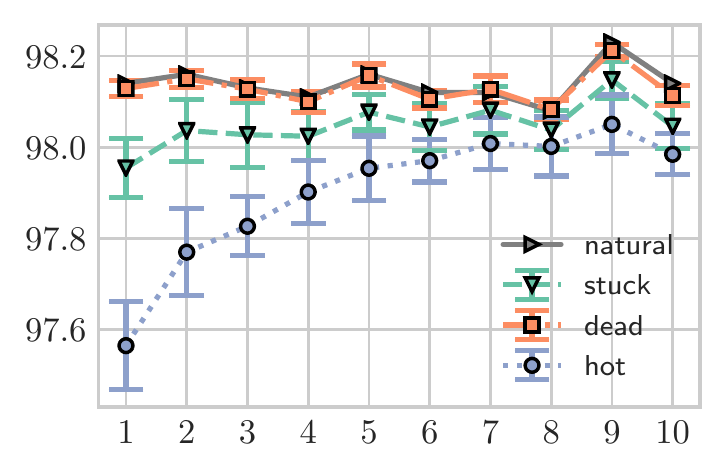}}\;
    \subfloat[Stuck\label{fig:transfer-cifar-stuck}]{%
        \includegraphics[width=0.30\textwidth]{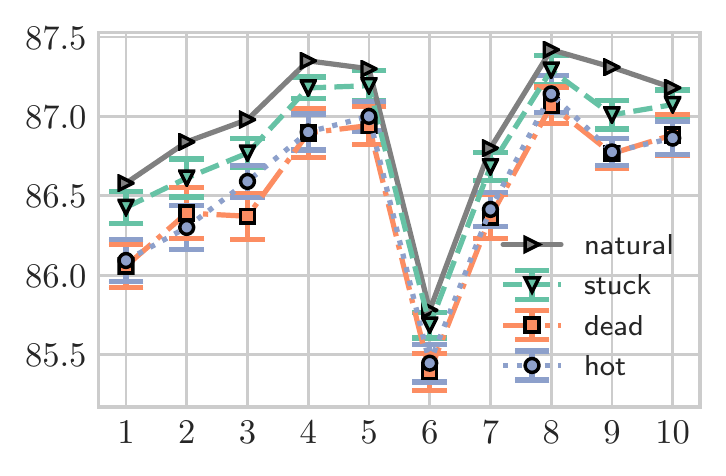}}
    
    \subfloat[Hot\label{fig:transfer-mnist-hot}]{%
        \includegraphics[width=0.30\textwidth]{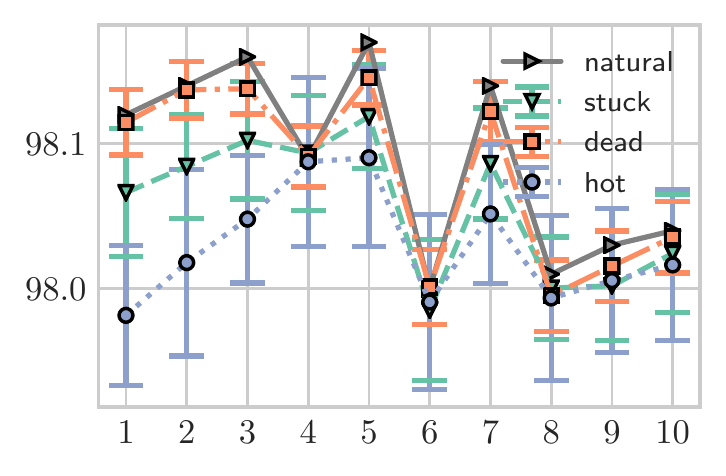}}\;
    \subfloat[Hot\label{fig:transfer-pixel-cifar-hot}]{%
        \includegraphics[width=0.30\textwidth]{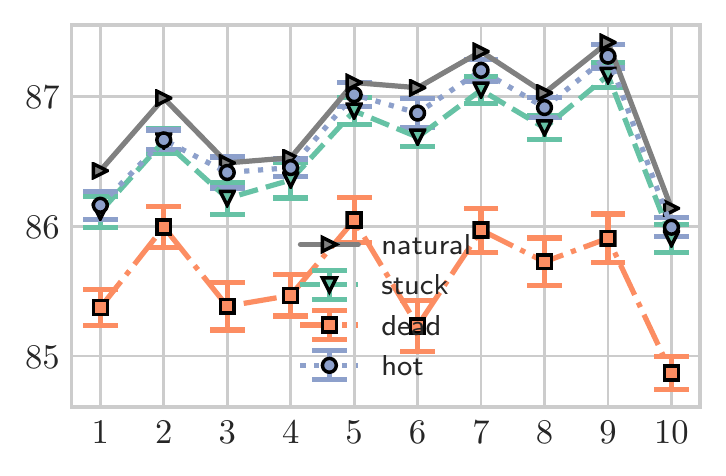}}
        
    \subfloat[Dead\label{fig:transfer-mnist-dead}]{%
        \includegraphics[width=0.30\textwidth]{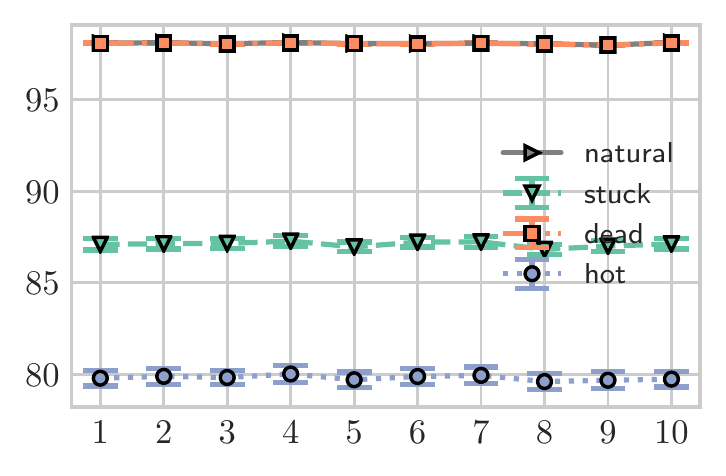}}\;
    \subfloat[Dead\label{fig:transfer-pixel-cifar-dead}]{%
        \includegraphics[width=0.30\textwidth]{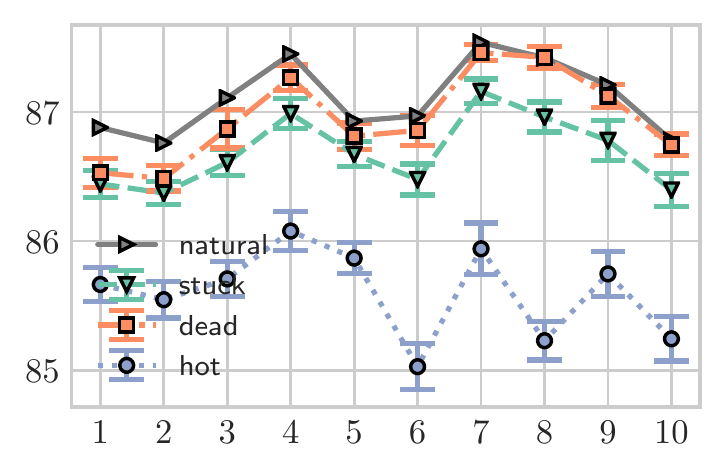}}
        
    \begin{minipage}{12cm}
    \caption{Transfer learning models trained on pixel variations for each pixel category. Results show tests on 2 px changes sampled 25 times for the MNIST and CIFAR-10 datasets. Test accuracy is shown as a function of the change in the number of training pixels from 1-10.} 
    \end{minipage}
    \label{fig:transfer-pixel}
\end{figure}

\begin{figure}[H]
    \centering
    \textbf{MNIST\hspace{12em}CIFAR-10}\par\medskip
    
    \subfloat[\label{fig:transfer-mnist-noise}]{%
        \includegraphics[width=0.30\textwidth]{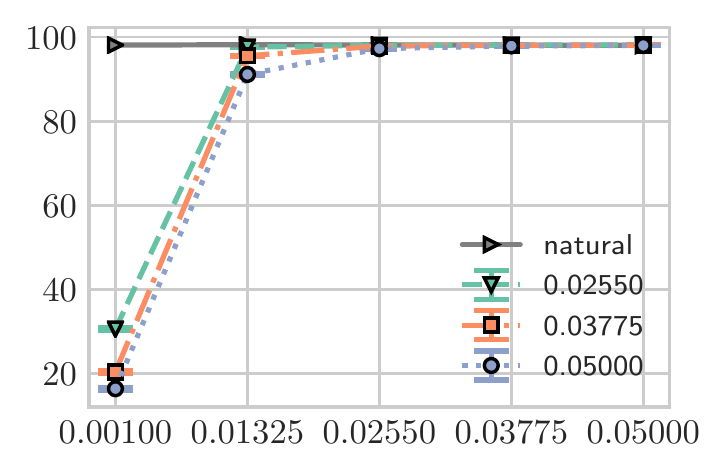}}\;
    \subfloat[\label{fig:transfer-cifar-noise}]{%
        \includegraphics[width=0.30\textwidth]{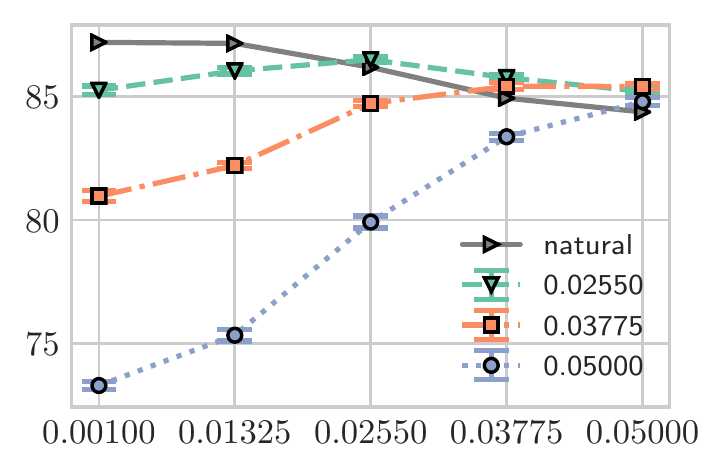}}
        
    \begin{minipage}{12cm}
    \caption{Transfer learning models trained on variations of noise sampled from a normal distribution. Results show tests on various noise levels for the MNIST and CIFAR-10 datasets. Test accuracy is shown as a function of the variance in noise.} 
    \end{minipage}
    \label{fig:transfer-noise}
\end{figure}

\begin{figure}[H]
    \centering
    \textbf{MNIST\hspace{12em}CIFAR-10}\par\medskip
    
    \subfloat[\label{fig:transfer-mnist-blur}]{%
        \includegraphics[width=0.30\textwidth]{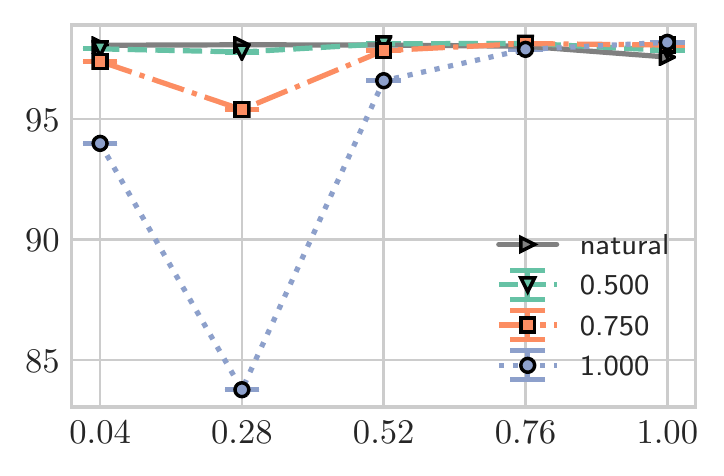}}\;
    \subfloat[\label{fig:transfer-cifar-blur}]{%
        \includegraphics[width=0.30\textwidth]{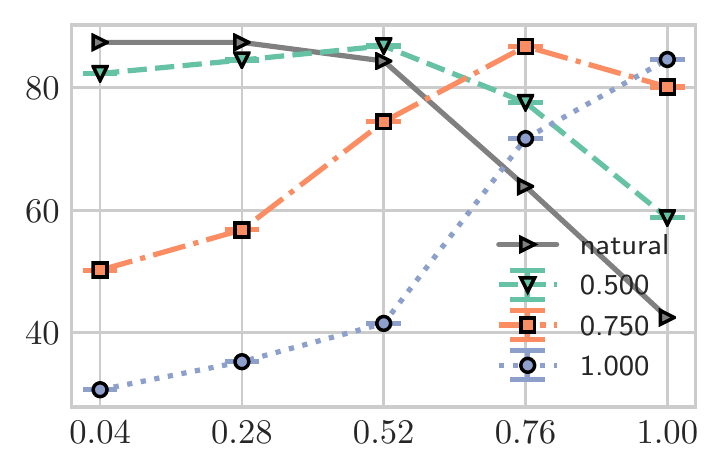}}
        
    \begin{minipage}{12cm}
    \caption{Impact of blurry input images on MNIST and CIFAR-10 models fine-tuned with transfer learning. Test accuracy is shown as a function of the variance in blur.} 
    \end{minipage}
    \label{fig:transfer-blur}
\end{figure}

\end{document}